# Extracting Semantic Concepts and Relations from Scientific Publications by Using Deep Learning

Fatima N. AL-Aswadi[1,2], Huah Yong Chan[1], and Keng Hoon Gan[1]

[1]School of Computer Sciences, Universiti Sains Malaysia, 11800 Gelugor, Pulau Pinang, Malaysia
[2]Faculty of Computer Sciences and Engineering, Hodeidah University, Hodeidah, Yemen

## Abstract

With the large volume of unstructured data that increases constantly on the web, the motivation of representing the knowledge in this data in the machine-understandable form is increased. Ontology is one of the major cornerstones of representing the information in a more meaningful way on the semantic Web. The current ontology repositories are quite limited either for their scope or for currentness. In addition, the current ontology extraction systems have many shortcomings and drawbacks, such as using a small dataset, depending on a large amount predefined patterns to extract semantic relations, and extracting a very few types of relations. The aim of this paper is to introduce a proposal of automatically extracting semantic concepts and relations from scientific publications. This paper suggests new types of semantic relations and points out of using deep learning (DL) models for semantic relation extraction.

## 1 INTRODUCTION

The substantial growth of unstructured data makes many applications of this data, such as information retrieval, information extraction or any other applications a hard and laborious task. This data contains much useful knowledge. Unfortunately, this knowledge is not in the machine-understandable form, it is just in human understandable form [1, 2]. Therefore, constructing the ontologies is considered an important task to make this data in the machine-understandable form as well as human understandable form. The ontology term is a data model to represent a set of concepts and the relations among those concepts within a domain [1].

There are many existing ontologies repositories or tools that are constructed or seek to construct ontologies either manually, semi-automatically or automatically. For example, WordNet which is considered one of the oldest and most popular ontology repositories. It is a high precise resource that was manually constructed by linguists. However, the progress of WordNet is quite slow comparing with streaming data across the web, as well as it lacks many modern terms, such as *cloud computing*, *deep learning* or even *netbook* [3].

Another example of ontology repositories is YAGO (Yet Another Great Ontology) [4], it is an ontology that built on top of both WordNet and Wikipedia. YAGO use the Wikipedia category pages rather than using information extraction methods to leverage the knowledge of Wikipedia. However, Wikipedia categories are often quite fuzzy and irregular this is considered one of the disadvantages of this repository [3] (it does not follow the expected pattern and it is open edit for anyone). Also, YAGO uses structured data for building the ontology which may result that the space may be wasted if not all arguments of n-array facts are known. In addition, YAGO is relatively little help if WordNet neither contains some of the related concepts [3].

There are many other ontology repositories that their ontologies were extracted from structured contents of Wikipedia pages such as *Freebase*, *BabelNet*, and *DBpedia*.

On the other hand, there are many of ontology extraction tools that try to extract and construct the ontology either semi-automatically or automatically, such as Text-to-Onto [5], SYNDIKATE (SYnthesis of DIstributed Knowledge Acquired from TExts) [6, 7], CRCTOL (Concept-Relation-Concept Tuple based Ontology Learning) [8], and ProMine [9].

Some of them using structured or semi-structured data as input to extract and construct the ontologies such as ProMine and Text-to-Onto, while others using unstructured data to extract the ontologies such as SYNDIKATE and CRCTOL. However, most of the existing ontology extraction tools have many shortcomings and drawbacks. For example, some of them depend on human intervention in the whole of their tasks such as Text-to-Onto. In addition, most of them such as Text-to-Onto and CRCTOL depend on predefined templates for relation extraction that lead to very low recall results. Moreover, some of these tools use small dataset such as Text-to-Onto which used only 21 web articles as the input dataset.

Nowadays, many research that tries to extract ontologies from scientific publications have begun to emerge, such as in [10, 11]. The [10] study is association rules-based approach for enriching the domain ontology rather than extracting new domain ontology. This study depends partially on lexical similarity measures, but in many cases, there is no correlation between the lexical similarity of concept names and the semantic concept similarity because of the high complexity of language or the uncoordinated ontology development. For example of this shortcoming, the concepts pair (table, stable) has lexical similarity while there is not semantically matching.

In the [12] study, the authors defined NTNU system that aim to extract the keyphrases and relations from scientific publications using multiple conditional random fields (CRFs), this study has many limitations and shortcomings as the author stated themselves. One main limitation of these limitations is that this study extract only two types of relations they are synonym and hyponym. In addition, the authors stated that their multiple CRF models with the help of rules have improved the performance on the development set, but the performance was worse on the testing set.

This paper gives a proposal of automatically extracting the semantic concepts and relations from scientific publications by using DL.

## 2 PROPOSED WORK

The main two tasks of the ontology constructing process are extracting the concepts, and extracting and mapping the relationship between these concepts. The high degree of precision and recall in these extracted relationships means the highest degree of precision and reliability of the constructed ontology.

The three main drawbacks and shortcomings in the most existing ontology construction research are:
1. Most of them depend on large amount predefined patterns such as in [3, 8, 10].
2. Many of them use a small dataset for constructing the ontologies such as in [5, 8]
3. Most of them extract very limited relations almost do not exceed *synonym, hyponym, hypernym, meronyms,* and/or *holonyms* relations such as in [3, 8, 9, 11].

Our proposed work aims to handle the above shortcomings by suggesting six more relation types for handling the third shortcoming and by using DL techniques for handling first and second shortcomings. That is because DL can handle a large amount of data in an efficient and effective way as well as because using predefined patterns can may give a reasonable precision, but a very low recall because that any relation is not within the predefined patterns cannot be detected. While DL is based on the deep learning fundament.

DL is a branch of neural networks (NNs), the difference between traditional NN and DL is in their architectures. NN have shallow architectures (one hidden layer); while DL has deep architectures (more than one hidden layer) and every hidden layer learns a new extracted features (concepts or relations) from the previous layer. That means every next layer gets more accurate results than the previous layer. The shallow architectures can effectively solve many simple, well-constrained or defined problems, but their modelling and representational power are limited [13]. Hence, for more complicated real-world applications such as human speech and natural language understanding, where we do not have enough predefined patterns or where we do not have a clear perception of problems, the deep architectures have more abilities when dealing with these complicated problems rather than shallow architectures [13]. As well as DL can handle a large amount of data in an effective and efficient way.

Table 1: Semantic relations

| Relation type | Example | Linguistic relation |
|---|---|---|
| Equal | data, information | Synonyms |
| Is_A | bubble sort, sorting algorithm | Hyponyms Hypernyms |
| Has_A | algorithm, performance | Holonyms |
| Different_of | plant, plant | Homonyms |
| Part_of | room, house | Meronyms |
| Used_to - Used_by | technology, waste food technology, human | Usage |
| Result_of | reliable ontology, precision relation | Result |
| Compared_to | bubble sort, merge sort | Comparison |
| Use_A - | image classification, machine learning | Model |
| Depend_On | performance, data size | Dependence |

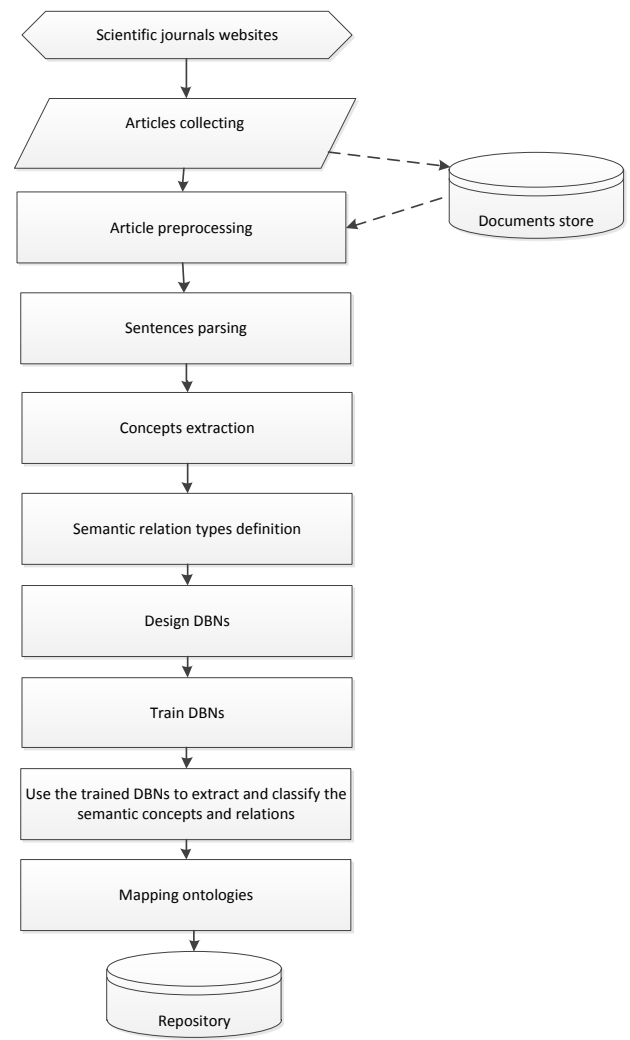

Figure 1: Workflow to extract semantic concepts and relations from scientific publications.

Table 1 shows the main types of semantic relations that this proposed work suggests extracting it from scientific publications, and it shows also the examples of these semantic relations. While figure 1 shows the workflow of the proposed work to extract the semantic concepts and relations from scientific publications.

In our previous work [14], we discussed and presented in details the literature review about ontology construction, OL, DL, and the DL for OL. In this paper, we introduce a proposal of extracting semantic concepts and relations by using Deep Belief Networks (DBNs). DBNs is one of the milestone models on the DL. Based on the current literature of DL field, it is observed that the DBN model is more appropriate for the tasks at suggested work in this paper.

In this work, DBNs is used to classify the extracted concepts and to extract the relations between concepts. Also, it is used to classify the extracted semantic relations under the main relations types that were defined in table 1.

After preprocessing the text and extracting the concepts from the text by using n-gram and other concept extraction methods, the terms and tags bags are built in binary representation. Then the system assigning the part of speech (POS) and syntactic tags to each individual term in binary representation. This combining between aims to build the feature vectors (training file for DBN).

After building appropriate DBN and training it, the trained DBN can be used to classify the concepts and to extract the relations. The concept classification by using DBN will be through two processes. First one is *detection process*, in this process DBN has only two target outputs: "yes" and "no" where "yes" means that $c_1 \subseteq c_2$ while "no" means $c_1 \nsubseteq c_2$ for each input vector ($c_i$ refers to concept *i*). The second process is the *classification process*, in this process the second and the third up the level of the *k* most relevant concepts are used as the DBN target outputs. The *k* most relevant concepts are identified by using some measures such as TF/IDF (term frequency / inverse document frequency). The same processes are done for semantic relations detection and classification except that the target outputs of the classification process are the relations identified in table 1.

It is worth mentioning that different DBNs can be used for different processes. Despite all the promising results of using DL for ontology construction, but the main problem of using the DL for ontology construction is to build an appropriate deep network and pick up the right method according to the particular task. That is what should be regarded in the current and future researches of ontology construction.

## 3 Conclusion

In this paper, we introduced new types of semantic relations and presented a proposal for extracting semantic concepts and relations automatically from scientific publications by using DL. This proposal aims to address the three main shortcomings and drawbacks in current ontology extraction systems as it pointed out above in the proposed work.